\documentclass{article} %
\usepackage[dvipsnames]{xcolor}
\usepackage{iclr,times}

\usepackage{amsmath,amsfonts,bm}

\def\eqref#1{equation~\ref{#1}}

\def\1{\bm{1}}

\DeclareMathAlphabet{\mathsfit}{\encodingdefault}{\sfdefault}{m}{sl}
\SetMathAlphabet{\mathsfit}{bold}{\encodingdefault}{\sfdefault}{bx}{n}

\PassOptionsToPackage{hyphens}{url}\usepackage{hyperref}
\usepackage{cleveref}
\usepackage{url}
\usepackage{graphicx}
\usepackage{subcaption}
\usepackage{booktabs}
\usepackage{wrapfig}
\usepackage{float}
\usepackage{algorithm}
\usepackage[noend]{algpseudocode}

\definecolor{cornflowerblue}{rgb}{0.39, 0.58, 0.93}
\hypersetup{
    colorlinks=true,
    linkcolor=cornflowerblue,
    filecolor=magenta,      
    urlcolor=teal,
    citecolor=cornflowerblue,%
    pdfpagemode=FullScreen,
    }

\title{Efficient Vision-Language Models by Summarizing Visual Tokens into Compact Registers}

\author{Yuxin Wen$^1$\textsuperscript{\thanks{Work done during an internship at Apple.}}\\
\And
Qingqing Cao$^2$
\And Qichen Fu$^2$
\And Sachin Mehta$^3$\textsuperscript{\thanks{Work done while working at Apple.}}
\And Mahyar Najibi$^2$ \AND
\vspace*{-0.8cm}\\
$^1$University of Maryland \quad \quad $^2$Apple \quad \quad $^3$Meta
}

\iclrfinalcopy %
\begin{document}

\maketitle
\begin{abstract}
Recent advancements in vision-language models (VLMs) have expanded their potential for real-world applications, enabling these models to perform complex reasoning on images. 
In the widely used fully autoregressive transformer-based models like LLaVA, projected visual tokens are prepended to textual tokens. Oftentimes, visual tokens are significantly more than prompt tokens, resulting in increased computational overhead during both training and inference. In this paper, we propose \textbf{Vi}sual \textbf{C}ompact \textbf{To}ken \textbf{R}egisters (\texttt{Victor}), a method that reduces the number of visual tokens by summarizing them into a smaller set of register tokens. \texttt{Victor} adds a few learnable register tokens after the visual tokens and summarizes the visual information into these registers using the first few layers in the language tower of VLMs. After these few layers, all visual tokens are discarded, significantly improving computational efficiency for both training and inference. Notably, our method is easy to implement and requires a small number of new trainable parameters with minimal impact on model performance.
In our experiment, with merely $8$ visual registers—about $1\%$ of the original tokens—\texttt{Victor} shows less than a $4\%$ accuracy drop while reducing the total training time by $43\%$ and boosting the inference throughput by $3.3\times$.
\end{abstract}

\section{Introduction}
Vision-language models (VLMs) have attracted considerable attention for their capability to process visual and textual information, enabling various real-world applications, such as image captioning, visual question answering, and multimodal reasoning \citep{openai2gpt4v, liu2024visual}. For example, GPT-4V \citep{openai2gpt4v} demonstrates the potential of these models in helping visually impaired individuals to ``see'' the world through cell phone cameras.

Recent transformer-based vision-language models, such as LLaVA \citep{liu2024visual}, employ a pre-trained vision encoder as the model's ``eye'' to extract visual features and use a pre-trained language model as the ``brain'' to perform reasoning and text generation. This simple architecture is highly effective and requires only a small instruction-based fine-tuning dataset for achieving state-of-the-art results on standard benchmarks. The visual tower in VLMs decomposes high-resolution images into a large number of visual tokens, which are then concatenated with prompt tokens as an input to the language tower. This process significantly increases the computational cost due to the quadratic attention cost with respect to tokens. As an example, LLaVA-NeXT \citep{liu2024llavanext} uses $2,880$ tokens to represent a single image, which can be overly redundant in many scenarios. In contrast, the average text instruction length across all benchmarks used in LLaVA-NeXT has fewer than $70$ tokens, as shown in \Cref{subsec:bench_stats}. Therefore, to improve the efficiency of VLMs, reducing the number of visual tokens is essential. 

The recent state-of-the-art method, FastV \citep{chen2024image}, achieves this by dropping unimportant visual tokens. This approach is highly effective when reducing the number of tokens by up to half. However, the model's performance drops significantly when more than half of the tokens are removed. Moreover, FastV requires the retrieval of attention scores from the self-attention block. Efficient attention implementations, such as FlashAttention \citep{dao2022flashattention, dao2023flashattention}, do not support this feature because they avoid storing the entire attention score matrix to optimize memory usage and computational efficiency. Consequently, FastV relies on standard attention implementations to retrieve the attention scores, which limits its deployment scenarios. We also find that this constraint slows down the fine-tuning stage (\Cref{subsec:training_time}). 

Alternatively, another popular approach for token reduction involves using a transformer-based projection model to condense visual tokens into a smaller set of queries. Notable examples include Perceiver Resampler \citep{jaegle2021perceiver, alayrac2022flamingo, bai2023qwen} and Q-Former \citep{li2023blip}. These methods outperform FastV when the reduced set of visual tokens is significantly smaller than the original. However, they require many more trainable parameters.

\begin{figure}[t!]
    \centering
    \includegraphics[width=0.95\textwidth]{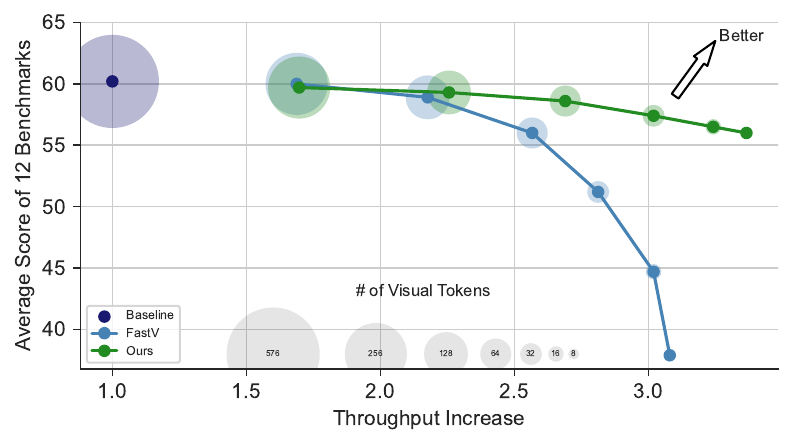}
    \caption{\textbf{Efficiency-Performance Trade-Off Curve.} We compare our proposed method, \texttt{Victor}, with the state-of-the-art method FastV. We report the normalized average score across $12$ benchmarks and the corresponding throughput increase relative to the original baseline model (details in \Cref{subsec:eval}). The size of the circles represents the number of visual tokens for each method, with larger circles representing more tokens. \texttt{Victor} establishes a more favorable Pareto frontier than FastV, demonstrating a significantly smaller performance drop as throughput increases.}
    \label{fig:teaser}
\end{figure}

In this paper, we introduce \textbf{Vi}sual \textbf{C}ompact \textbf{To}ken \textbf{R}egisters (\texttt{Victor}), a simple yet effective early visual token summarization method that provides a better efficiency-quality trade-off compared to the state-of-the-art methods. For visual tokens, we observe that they often contain redundant information, with many tokens being highly similar, as discussed in \Cref{subsec:motivation}. To address this, our approach leverages the LLM to summarize these visual tokens into compact registers. \texttt{Victor} begins by appending a small set of register tokens to the visual tokens and uses the initial $k$ layers of LLM to summarize visual information into these registers. Notably, during training, no additional loss function or operation is required to explicitly force summarizing the visual information into the registers. However, empirical results show that the language model naturally uses these tokens to store visual information. As a result, starting at layer $k$, all visual tokens are discarded, leaving only the summarized registers and textual tokens for efficient inference in the subsequent layers.

Our method does not rely on attention scores and can be seamlessly integrated with efficient attention implementations. Moreover, unlike approaches like Perceiver Resampler which deploy a separate transformer, \texttt{Victor} leverages the powerful language model itself to perform this task, introducing only $1.78$M additional parameters, accounting for only $0.03\%$ of the total model size, and achieves a significantly better performance.

As illustrated in \Cref{fig:teaser}, \texttt{Victor} establishes a more favorable Pareto frontier than the state-of-the-art method, particularly exhibiting a noticeably smaller performance drop as throughput increases. For instance, when the number of visual registers is set to $8$—approximately only $1\%$ of the original visual tokens—our model experiences a performance drop of less than $4\%$ while reducing total training time by $43\%$ and achieving a $3.36\times$ increase in inference throughput. Our extensive experiments demonstrate the effectiveness of \texttt{Victor} in balancing both efficiency and performance, making it a promising solution for reducing visual tokens in vision-language models.

\section{Related Work}
\subsection{Vision-Language Models}

Modern vision-language models typically combine a pre-trained image encoder with a large language model to handle multimodal data \citep{li2023blip}. One popular approach, often referred to as the autoregressive or LLaVA-style model \citep{li2023blip, liu2024visual}, directly projects visual features into the input embedding space of the language model, treating these features as part of the input tokens. However, in this design, the number of visual tokens is large and often exceeds the number of textual tokens, leading to inefficiencies. Another common approach is cross-attention-based fusion \citep{alayrac2022flamingo}, where added cross-attention blocks inside of the language transformer layers allow textual tokens to attend to visual tokens. More recently, early-fusion models like Fuyu \citep{fuyu-8b}, MoMA \citep{lin2024moma}, and Chameleon \citep{team2024chameleon} use a unified transformer that processes raw textual tokens and visual patches simultaneously. Additionally, a key component of modern vision-language models recipe is instruction fine-tuning \citep{Dai2023InstructBLIPTG, zhu2023minigpt, liu2024improved, liu2024visual, singla2024pixels}, which enables the model to function as a typical chatbot while also processing images, even with a small synthetic fine-tuning dataset. In this paper, we focus on LLaVA-style models.

\subsection{Visual Token Reduction}

To improve the efficiency of vision or vision-language models, pruning or distilling visual tokens has been widely studied. \citet{rao2021dynamicvit} introduce DynamicViT, which uses a small module to predict the importance of each visual token, dropping less important ones to enhance efficiency. Similarly, EViT \citep{liang2022not} retains important tokens and fuses the less important ones within the model, using attention scores from the class token to the visual tokens. Further, PuMer \citep{cao2023pumer} reduces the number of both textual and visual tokens by progressively pruning and merging them throughout the cross-modal encoder. Another interesting approach by \citet{saifullah2023seeing} involves discretizing visual features into textual tokens to reduce dimensionality. For more recent vision-language models, Perceiver Resampler \citep{jaegle2021perceiver, alayrac2022flamingo, bai2023qwen} and Q-Former \citep{li2023blip} are commonly used to pool visual tokens into a smaller set of queries using a transformer-based model. Additionally, in the FastV paper \citep{chen2024image}, the authors observe that in LLaVA-style models, the attention from textual tokens to visual tokens significantly diminishes after the first few layers, with the attentiveness declining close to zero after $10\%$ of the transformer layers. Intuitively, their proposed state-of-the-art method drops the unimportant visual tokens accordingly after the first few layers.

\subsection{Visual Registers}

\citet{burtsev2020memory} first introduce memory tokens, which function similarly to registers. These tokens are used to store global information, enabling the model to effectively handle long-context tasks. \citet{darcet2023vision} apply the idea of registers to ViTs. In their work, the authors observe that vision transformer models implicitly use low-information tokens to store global information for internal computations. However, this leads to abnormally high norms for these tokens, making it difficult to interpret the attention maps. Therefore, to address this, they introduce additional register tokens at the end of the sequence to handle this task. This approach not only improves the interpretability of attention maps but also boosts model performance. In our work, we show that these register tokens also enhance the performance of vision-language models, as demonstrated in \Cref{subsec:effect_of_register}. However, our primary focus in this paper is on using these registers for information distillation, enabling the model to condense visual information into the registers to improve efficiency.

\section{Method}
\begin{figure}[t!]
    \centering
    \includegraphics[width=\textwidth]{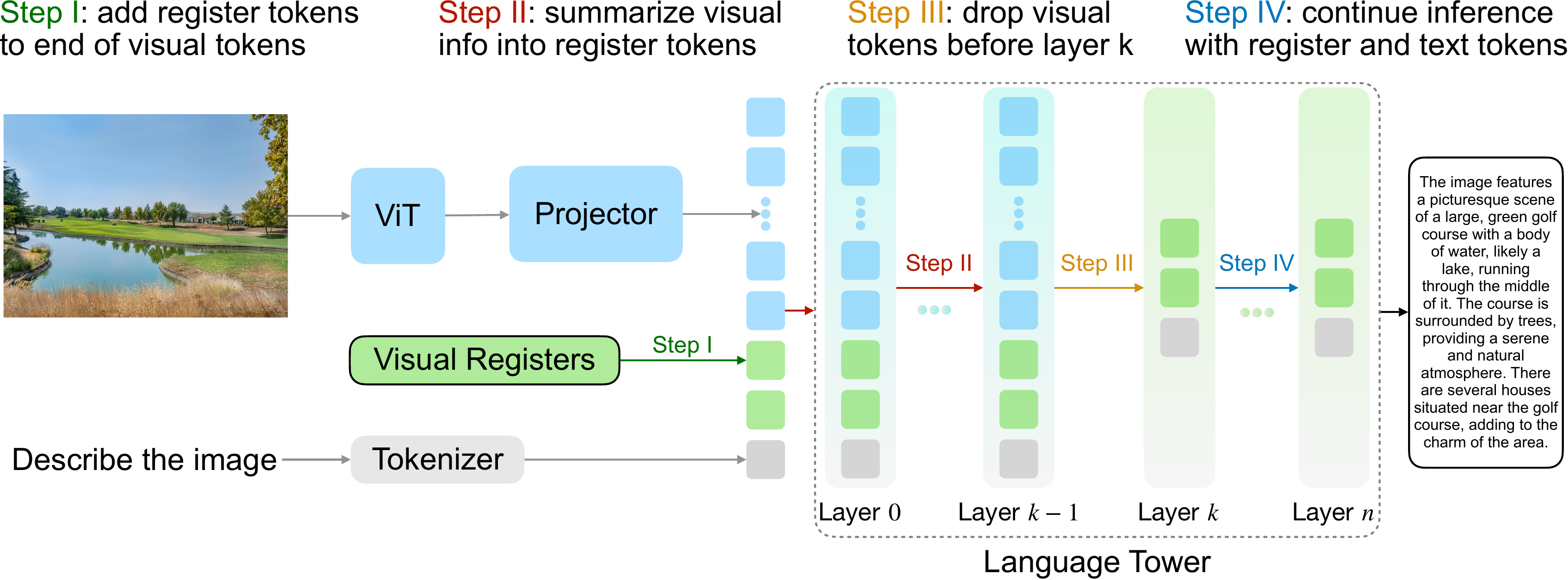}
    \caption{\textbf{Method Overview.} \texttt{Victor} is a simple yet effective method for enhancing the efficiency of vision-language models. The process involves four key steps based on the LLaVA-style model: (I) appending learned visual register tokens after the visual tokens, where the number of visual registers is much smaller than the number of the visual tokens, (II) using the first $k$ layers of the language tower to summarize visual information into the visual registers, (III) discarding all visual tokens before layer $k$, and (IV) starting from layer $k$, the model performs efficient inference using only the visual registers and textual tokens with significantly reduced sequence length.}
    \label{fig:method_overview}
\end{figure}

\subsection{Motivation}
\label{subsec:motivation}

\begin{wrapfigure}{t!}{0.5\textwidth}
  \vspace{-0.1cm}
  \begin{center}
    \includegraphics[width=0.49\textwidth]{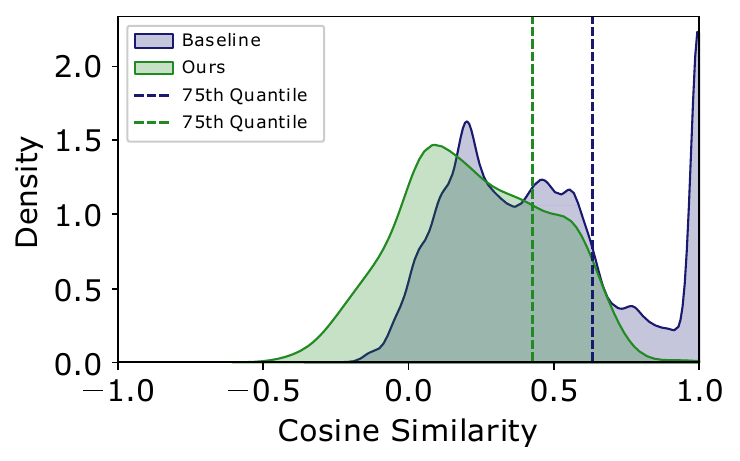}
  \end{center}
  \vspace{-0.3cm}
  \caption{\textbf{Token Similarities.}}
  \label{fig:token_sim}
  \vspace{-0.5cm}
\end{wrapfigure}

We start with a key observation: many visual tokens exhibit significant redundancy. To demonstrate this, we calculated the cosine similarities between visual tokens generated by vision towers in VLMs. As shown in \Cref{fig:token_sim}, these similarities tend to cluster around $1$, indicating a high degree of similarity among the visual tokens. This suggests that compressing the visual tokens into a smaller set would result in a minimal information loss. To achieve this, we append a set of learnable register tokens to the visual tokens and leverage the language model to summarize the visual information into these registers. Rather than using a separate model, such as the Perceiver Resampler, we utilize the more powerful language model for this task, as it inherently understands which visual tokens are important and how to organize the information. Consequently, as demonstrated in \Cref{fig:token_sim}, our compact visual registers show reduced redundancy compared to the baseline visual tokens.

Furthermore, based on observations from FastV \citep{chen2024image}, textual tokens in the language model primarily attend to visual tokens in the early transformer layers, with attention scores to visual tokens dropping to nearly zero after these layers. This suggests the language model requires only a few layers to process the visual information. Therefore, instead of using the entire model, we employ only the first few transformer layers to summarize the visual tokens into the register tokens. After summarization, the visual tokens are discarded, improving model efficiency. An overview of our method is provided in \Cref{fig:method_overview}\footnote{Image credit: \url{https://unsplash.com/photos/green-grass-field-near-lake-under-blue-sky-during-daytime-d3Jf3avtXSg}}.

\subsection{Victor}
We now formally introduce our method: \texttt{Victor} (\textbf{Vi}sual \textbf{C}ompact \textbf{To}ken \textbf{R}egisters). A LLaVA-style vision-language model consists of three main components: (1) the image tower $\mathcal{I}$, which is a pre-trained vision model, such as the CLIP image encoder \citep{radford2021learning}; (2) the language tower $\mathcal{T}$, a pre-trained LLM, such as LLaMA \citep{touvron2023llama}; and (3) a projector $\mathcal{P}$ that bridges the two, mapping the image features into the input embedding space of the language model. Given an image $x_{\text{img}}$, we first extract its features using the image tower $\mathcal{I}$ and produce a set of projected visual tokens $x_V = \{x_V^0, x_V^1, \ldots, x_V^{N-1}\}$ from the projector $\mathcal{P}$.

\begin{algorithm}[!t]
\caption{\texttt{Victor} - \textbf{Vi}sual \textbf{C}ompact \textbf{To}ken \textbf{R}egisters}
\label{alg:algorithm}
\begin{algorithmic}[1]
\State \textbf{input} Projected Visual Tokens: $x_V = \{x_V^0, x_V^1, \ldots, x_V^{N-1}\}$, Textual Input Tokens: $x_T = \{x_T^0, x_T^1, \ldots, x_T^{L-1}\}$, Visual Registers: $x_R = \{x_R^0, x_R^1, \ldots, x_R^{M-1}\}$, Language Tower: $\mathcal{T}$, Number of Layers: $n$, Drop Layer Index: $k$
\State Form input $x = [x_V; x_R; x_T]$ \Comment{Add visual registers
to end of visual tokens}
\For{each layer $i = 0$ to $n$}
    \If{$i == k$}
        \State $x = x[N:]$ \Comment{Drop visual tokens before layer $k$}
    \EndIf
    \State $x = \mathcal{T}_{i}(x)$
\EndFor
\State \Return $x$
\end{algorithmic}
\end{algorithm}

As described in \Cref{alg:algorithm}, for \texttt{Victor}, we additionally introduce a set of learnable visual registers $x_R = \{x_R^0, x_R^1, \ldots, x_R^{M-1}\}$, where $M$ is a hyperparameter controlling the number of visual registers. A smaller $M$ results in a more efficient model, and usually $M \ll N$. We then concatenate the projected visual tokens, visual registers, and textual tokens to form the input: $x = [x_V; x_R; x_T]$. This input is processed through the language tower for the first $k$ layers. At the start of layer $k$, all visual tokens are discarded, and the model continues with the truncated hidden states for the remaining layers.

During training, we do not explicitly force the language model to summarize the visual information into the visual registers, but we empirically observe that it does so implicitly. In \Cref{subsec:analysis}, we provide an empirical analysis showing that the visual registers effectively summarize important information from the visual tokens. Moreover, because \texttt{Victor} leverages the language model itself for this summarization, rather than relying on an external model, and the language model is both powerful and knowing at identifying the most useful image features, our method experiences minimal performance drop compared to approaches like Perceiver Resampler while requiring much fewer additional model parameters.

In practice, we typically set $k$ to $3$, which is approximately $10\%$ of the language tower. This means the language tower processes the full-length hidden states for only the first $10\%$ of its layers. For the remaining $90\%$ layers, it operates on a significantly shorter context, thereby improving model efficiency. FastV \citep{chen2024image}, a state-of-the-art method, follows a similar idea and drops unimportant tokens in the early layers and achieves a comparable theoretical FLOPs reduction. However, we find that since FastV relies on attention scores to determine which tokens to drop, it cannot utilize efficient attention implementations like FlashAttention \citep{dao2022flashattention, dao2023flashattention} or PyTorch High-Performance Scaled Dot Product Attention (SDPA). Consequently, FastV delivers less improvement in throughput than \texttt{Victor} when applying the same token-drop ratio in practice, and also empirically experiences a greater performance degradation in high token-drop ratio regimes.
\section{Experiments}
\subsection{Experimental Setup}
\label{subsec:setup}
In this paper, we primarily follow the setting of the open-sourced LLaVA-v1.5 \citep{liu2024improved}. The training consists of two main stages: pre-training and instruction fine-tuning.

\textbf{Pre-trained Models.} For the image tower, we use the pre-trained OpenAI CLIP ViT-Large model \citep{radford2021learning}, and for the text tower, we use the Vicuna-7B-v1.5 model \citep{zheng2024judging}, an instruction fine-tuned version of LLaMA-2-7B \citep{touvron2023llama}. Additionally, we present results using different language towers in \Cref{subsec:diff_model}, including Vicuna-13B-v1.5 \citep{zheng2024judging}, Meta-Llama-3-8B-Instruct \citep{dubey2024llama}, Mistral-7B-Instruct-v0.2 \citep{jiang2023mistral}, and Qwen2-7B-Instruct \citep{yang2024qwen2}.

\textbf{Datasets.} For pre-training, we use the LLaVA CC3M pre-training dataset \citep{liu2024visual}\footnote{\url{https://huggingface.co/datasets/liuhaotian/LLaVA-CC3M-Pretrain-595K}}, a subset of $595$K images from the CC3M dataset \citep{sharma2018conceptual}. For instruction fine-tuning, we use LLaVA-v1.5-mix$665$K\footnote{\url{https://huggingface.co/datasets/liuhaotian/LLaVA-Instruct-150K/blob/main/llava_v1_5_mix665k.json}}, a mixed dataset comprising COCO 2017 \citep{lin2014microsoft}, GQA \citep{hudson2019gqa}, OCR-VQA \citep{mishraICDAR19}, TextVQA \citep{singh2019towards}, and VisualGenome \citep{krishna2017visual}.

\textbf{Implementation.} 
We follow the hyperparameters used by \citet{liu2024visual}. During pre-training, we freeze both the image and text towers, training only the projector and registers. We use the AdamW optimizer \citep{loshchilov2017decoupled} with a learning rate of $0.0001$ and no weight decay for one epoch. In the fine-tuning stage, we unfreeze the text tower while keeping the image tower frozen. As in pre-training, we use the AdamW optimizer with a learning rate of $0.00002$ and no weight decay for one epoch.

\vspace{-0.1cm}
\subsection{Evaluation}
\label{subsec:eval}
We use LMMs-Eval \citep{lmms_eval2024} and run evaluations over the $11$ tasks reported in the LLaVA-v1.5~\citep{liu2024improved} (\textit{i.e.} VQAv2 \citep{goyal2017making}, GQA \citep{hudson2019gqa}, ScienceQA \citep{lu2022learn}, TextVQA \citep{singh2019towards}, VizWiz-VQA \citep{gurari2018vizwiz}, POPE \citep{li2023evaluating}, MME \citep{yin2023survey}, MMBench \citep{liu2023mmbench}, SEED-Bench \citep{li2023seed}, LLaVA-Bench-in-the-Wild \citep{liu2024visual}, MM-Vet \citep{yu2023mmvet}), supplemented by the MMMU task \citep{yue2024mmmu}. These benchmarks provide a comprehensive assessment of models' multi-modal reasoning capabilities, encompassing academic-task-oriented and instruction-following tasks. For simplicity, we primarily report the average of normalized benchmark scores. Specifically, for MME, the score is divided by $2,000$, which represents the full score, as its metric is calculated by summing the accuracies of individual subtasks.

We evaluate efficiency by measuring the increase in throughput with the KV-cache enabled \citep{pope2023efficiently}. After gathering statistics from all $12$ benchmarks, presented in \Cref{subsec:bench_stats}, we evaluate two settings: 1) $2$-token generation and 2) $128$-token generation. The $2$-token generation simulates scenarios where questions expect a single word, as in GQA \citep{hudson2019gqa} and TextVQA \citep{singh2019towards}. In contrast, the $128$-token generation represents open-ended question scenarios, such as in LLaVA-Bench-in-the-Wild \citep{liu2024visual} and MM-Vet \citep{yu2023mmvet}. In both settings, we use a text prompt length of $64$ and a batch size of $16$. We choose a batch size of $16$ because it is the largest batch size that fits in memory. All training is conducted on $8$ NVIDIA A100 GPUs, with efficiency profiling performed on a single NVIDIA A100. By default, we set $k$ to $3$ and vary the number of final visual tokens to $256$, $128$, $64$, $32$, $16$, and $8$ for all methods to thoroughly assess the trade-off between efficiency and performance, where the number of visual tokens for the original LLaVA-v1.5 model is $576$.

We compare our method against two baselines: FastV \citep{chen2024image} and Perceiver Resampler \citep{jaegle2021perceiver}. FastV is the state-of-the-art token reduction method that filters out less important vision tokens based on attention scores, while the Perceiver Resampler is a compact, transformer-based model designed to condense input tokens into a smaller query set.

\section{Results}

\subsection{Throughput Increase}
\label{subsec:main_result}
\begin{figure}[!ht]
\centering
\begin{subfigure}[t]{0.49\textwidth}
    \includegraphics[width=\textwidth]{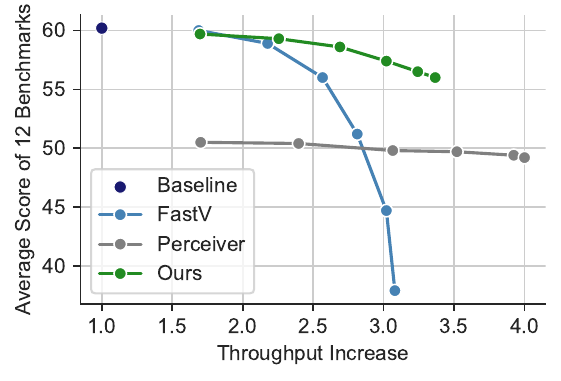}
    \caption{$2$-Token Generation}
    \label{fig:main_result_0}
\end{subfigure}
\hfill
\begin{subfigure}[t]{0.49\textwidth}
    \includegraphics[width=\textwidth]{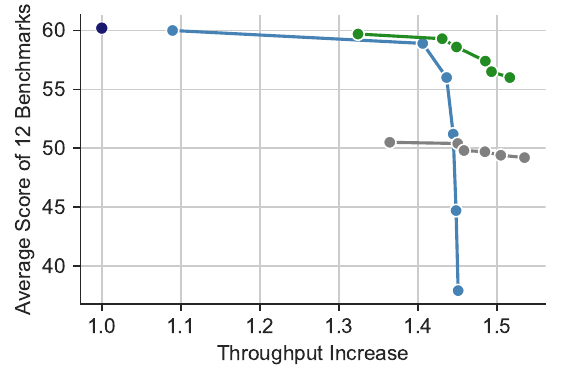}
    \caption{$128$-Token Generation}
    \label{fig:main_result_1}
\end{subfigure}
\caption{\textbf{Efficiency-Performance Trade-Off Curve.} We measure the relative throughput increase compared to the baseline model. The test covers two main scenarios: generating $2$ tokens and generating $128$ tokens. In both cases, the batch size is set to $16$, and the text prompt length is $64$ tokens. For all methods, we use $256$, $128$, $64$, $32$, $16$, and $8$ visual tokens to generate the line plot, and the number of visual tokens for the baseline method is $576$.}
\label{fig:main_result}
\end{figure}

We present the efficiency and performance trade-offs for both generation settings in \Cref{fig:main_result}, and the performance on individual benchmarks is provided in \Cref{subsec:individual_bench}. As shown, our method has a better Pareto frontier than FastV and the Perceiver Resampler in both scenarios.

Both \texttt{Victor} and FastV maintain minimal performance degradation when the throughput increases by approximately $1.5\times$ to $2\times$ and the number of tokens decreases from $576$ to $256$ or $128$. However, FastV’s performance declines rapidly beyond this point. In contrast, our method exhibits only a $4\%$ performance drop even when the number of visual tokens is reduced to $8$, which is roughly $1\%$ of the original visual token count. Additionally, with the same number of final visual tokens, our method has slightly higher theoretical FLOPs than FastV due to the inclusion of extra register tokens in the initial layers. However, in practice, our method achieves a greater increase in throughput compared to FastV. This is due to the fact that, in the layer where FastV performs filtering, the model is constrained to using the original attention mechanism to compute attention scores, as it cannot leverage more efficient attention implementations. In contrast, our method is compatible with a wide range of efficient attention implementations including those that do not support returning attention scores. As a result, \texttt{Victor} not only achieves better throughput and more effectively retains the accuracy but is also more adaptable across different devices than FastV.

\begin{wraptable}{r}{0.5\textwidth}
\centering
\caption{\textbf{Number of Extra Parameters for Different Methods.} The final number of visual tokens is $256$.}
\label{tab:num_extra_parameters}
\begin{tabular}{c|c}
\toprule
Method              & \# of Extra Parameters \\
\midrule
FastV               & $0$ ($0.00\%$)                     \\
Perceiver           & $252.86$M ($3.61\%$)               \\
Ours                & $1.78$M ($0.03\%$)   \\
\bottomrule
\end{tabular}
\vspace{-0.3cm}
\end{wraptable} 

In contrast, the Perceiver Resampler experiences a substantial performance drop of approximately $10\%$ relative to the original model, and performs significantly worse than \texttt{Victor}. Interestingly, its performance remains stable across different reduction ratios, consistent with the findings of \citet{laurenccon2024matters}. Despite this performance decline, the Perceiver Resampler achieves a much higher throughput increase than FastV and \texttt{Victor}. However, as shown in \Cref{tab:num_extra_parameters}, the Perceiver Resampler requires a substantially larger number of additional parameters—$252.86$M, representing $3.61\%$ of the total model—while our method adds only $1.78$M, approximately $0.03\%$ of the entire model.

\begin{wrapfigure}{!ht}{0.5\textwidth}
  \vspace{-0.3cm}
  \begin{center}
    \includegraphics[width=0.49\textwidth]{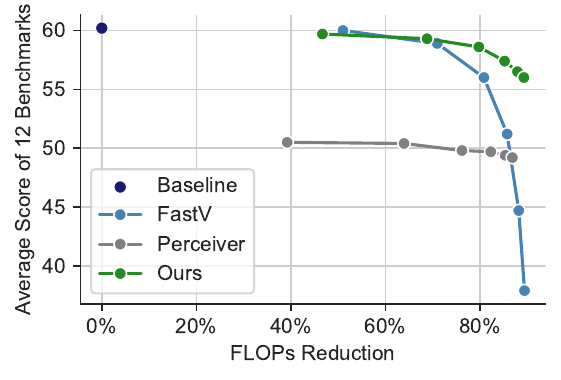}
  \end{center}
  \vspace{-0.5cm}
  \caption{\textbf{Performance vs. FLOPs Reduction.}}
  \label{fig:flops}
  \vspace{-0.5cm}
\end{wrapfigure}

\subsection{FLOPs Reduction}
We also report the theoretical FLOPs reduction of the methods, calculated using the FLOPs formula from \citet{chen2024image}. As demonstrated in \Cref{fig:flops}, while our method shows a slightly smaller FLOPs reduction due to the presence of additional register tokens at the start of the language tower, the overall reduction is comparable under the significantly higher token reduction rate. Although the Perceiver Resampler achieves a notable increase in throughput, its FLOPs reduction is substantially lower than that of FastV and \texttt{Victor}, primarily due to the additional transformer layers it employs.

\subsection{Training-Time Reduction}
\label{subsec:training_time}

\begin{figure}[!ht]
\centering
\begin{subfigure}[t]{0.49\textwidth}
    \includegraphics[width=\textwidth]{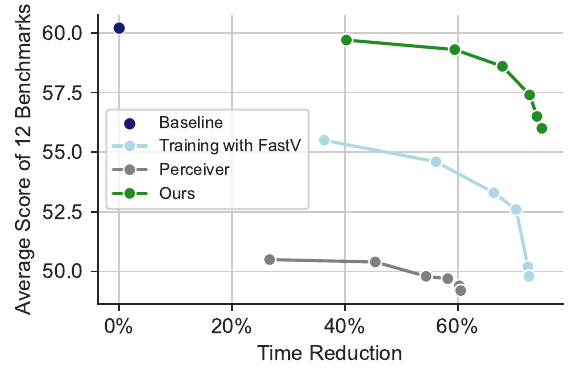}
    \caption{Pre-Training}
    \label{fig:training_time_result_0}
\end{subfigure}
\hfill
\begin{subfigure}[t]{0.49\textwidth}
    \includegraphics[width=\textwidth]{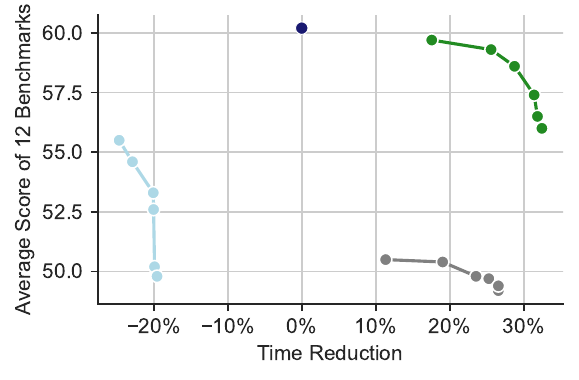}
    \caption{Fine-Tuning}
    \label{fig:training_time_result_1}
\end{subfigure}
\caption{\textbf{Performance vs. Training-Time Reduction.} We show total training-time reduction in \Cref{subsec:total_time}.}
\label{fig:training_time_result}
\end{figure}

\texttt{Victor} not only reduces inference costs but also lowers training costs. As indicated in \Cref{fig:training_time_result}, both Perceiver Resampler and \texttt{Victor} significantly reduce training time in both pre-training and fine-tuning stages, with the reduction being especially notable during pre-training due to the shorter text tokens. \texttt{Victor} achieves a greater overall time reduction. In contrast, training with FastV only reduces pre-training time and does not improve fine-tuning efficiency. This is because fine-tuning typically involves a large number of text tokens (often exceeding a thousand), and the use of a naive attention implementation in this phase introduces significant overhead, reducing training efficiency. Additionally, we observe that training with FastV does not match the performance of inference-time FastV. However, it exhibits slower benchmark performance decay as the number of visual tokens decreases and outperforms inference-time FastV when the number of visual tokens drops below $32$.

\begin{figure}[!ht]
\centering
\begin{subfigure}[t]{0.49\textwidth}
    \includegraphics[width=\textwidth]{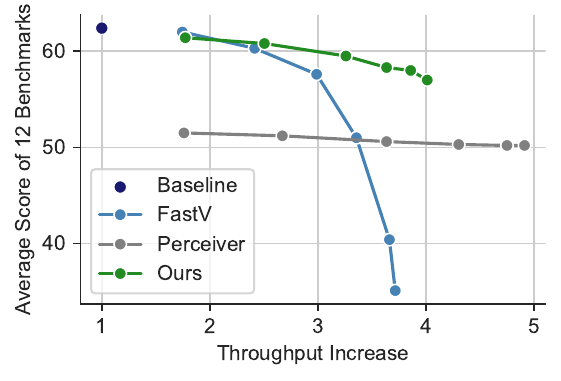}
    \caption{Vicuna-13B-v1.5}
    \label{fig:s0_diff_model_0}
\end{subfigure}
\hfill
\begin{subfigure}[t]{0.49\textwidth}
    \includegraphics[width=\textwidth]{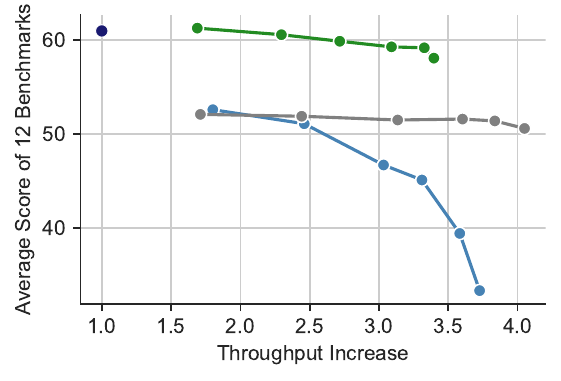}
    \caption{Meta-Llama-3-8B-Instruct}
    \label{fig:s0_diff_model_1}
\end{subfigure}
\hfill
\begin{subfigure}[t]{0.49\textwidth}
    \includegraphics[width=\textwidth]{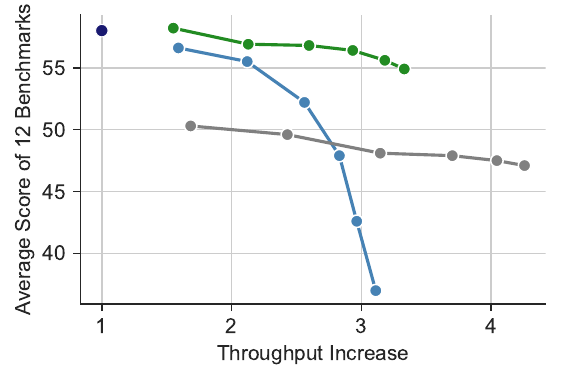}
    \caption{Mistral-7B-Instruct-v0.2}
    \label{fig:s0_diff_model_2}
\end{subfigure}
\hfill
\begin{subfigure}[t]{0.49\textwidth}
    \includegraphics[width=\textwidth]{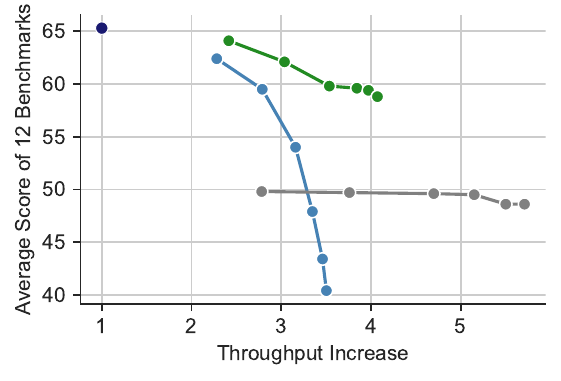}
    \caption{LLaVA-NeXT with Qwen2-7B-Instruct}
    \label{fig:s0_diff_model_3}
\end{subfigure}
\caption{\textbf{Efficiency-Performance Trade-Off Curve with Different Language Towers under 2-Token Generation.} Due to the space limit, we show the $128$-token generation scenario in \Cref{subsect:app_diff_models}. For the first $3$ models, we use $256$, $128$, $64$, $32$, $16$, and $8$ visual tokens to generate the line plot, and for LLaVA-NeXT, we use $512$, $256$, $128$, $64$, $32$, and $16$ visual tokens.}
\label{fig:diff_model}
\end{figure}

\subsection{Different Language Towers}
\label{subsec:diff_model}

We extensively evaluate the effectiveness of our method with different language towers. As shown in \Cref{fig:diff_model}, replacing the original Vicuna-7B-v1.5 language model with Vicuna-13B-v1.5 \citep{zheng2024judging}, Meta-Llama-3-8B-Instruct \citep{dubey2024llama}, and Mistral-7B-Instruct-v0.2 \citep{jiang2023mistral}, \texttt{Victor} remains highly effective and significantly outperforms the two baseline methods. For both Meta-Llama-3-8B-Instruct and Mistral-7B-Instruct-v0.2, \texttt{Victor} demonstrates minimal performance drop and a slow decay in performance as the number of visual tokens decreases. Notably, for these two models, when the number of visual tokens is reduced by half, the method shows no performance degradation at all.

We further demonstrate the performance of our method on a different vision-language model design: LLaVA-NeXT (LLaVA-v1.6) \citep{liu2024improved}. LLaVA-NeXT follows a similar architecture to LLaVA-v1.5 but increases the number of visual tokens from 576 to 2,880 by incorporating different aspect ratios, enhancing the model's capabilities. Additionally, LLaVA-NeXT utilizes Qwen2-7B-Instruct \citep{yang2024qwen2} as its language tower, benefiting from its extended context length. In our experiments, we reduce the number of visual tokens to $512$, $256$, $128$, $64$, $32$, and $16$. As indicated in \Cref{fig:s0_diff_model_3}, our method remains highly effective in the LLaVA-NeXT setting, consistently outperforming both FastV and the Perceiver Resampler.

\subsection{Different Layers to Drop Visual Tokens}
\label{subsec:drop_layer}
\begin{wrapfigure}{t!}{0.5\textwidth}
  \vspace{-0.7cm}
  \begin{center}
    \includegraphics[width=0.49\textwidth]{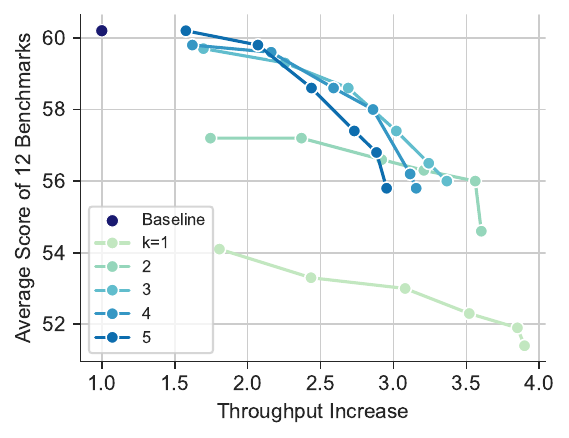}
  \end{center}
  \vspace{-0.5cm}
  \caption{\textbf{Ablation on Token Dropping Layers.}}
  \vspace{-0.3cm}
  \label{fig:diff_level}
\end{wrapfigure}
We show the results of the ablation study on which layer to drop the visual tokens (hyperparameter $k$) in \Cref{fig:diff_level}. In terms of throughput improvement, it is clear that the earlier we drop the visual tokens, the more efficient the model becomes. For lower-layer numbers, such as $k=1$ or $k=2$, the model's efficiency significantly increases, with throughput reaching nearly a $4\times$ improvement. However, this comes with a substantial performance drop, suggesting that one or two layers are likely insufficient for the summarization process. In contrast, when $k \geq 3$, the performance degradation is minimal, staying within a $5\%$ performance score loss. Notably, when $k=5$, with half of the visual tokens dropped, the model experiences no performance loss.

\subsection{Effect of Visual Registers on Regular VLMs}

\label{subsec:effect_of_register}

\begin{wrapfigure}{t!}{0.44\textwidth}
  \vspace{-1cm}
  \begin{center}
    \includegraphics[width=0.43\textwidth]{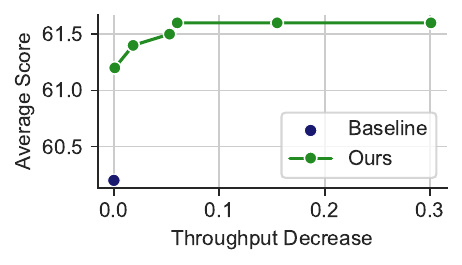}
  \end{center}
  \vspace{-0.5cm}
  \caption{\textbf{Results without Dropping Visual Tokens.} From left to right, we add $8$, $16$, $32$, $64$, $128$, and $256$ visual tokens \textit{resp}.}
  \label{fig:no_drop}
\end{wrapfigure}

\Cref{fig:no_drop} presents the results of not dropping the visual tokens and instead using visual registers as a means for the model to store useful information, similar to those proposed by \citet{darcet2023vision}. 

As reflected in \Cref{fig:no_drop}, there is a slight performance improvement over the baseline, but it is limited to around a $2\%$ increase. However, once the number of visual registers exceeds $64$, a further increase does not result in additional performance gains. On the other hand, adding more visual tokens leads to a decrease in throughput. Interestingly, adding just $8$ visual tokens offers a minimal throughput reduction while still providing a $1\%$ performance boost, making it almost a ``free lunch'' for visual-language models.

\subsection{Analysis}
\label{subsec:analysis}
\begin{figure}[!ht]
    \centering
    \includegraphics[width=0.95\textwidth]{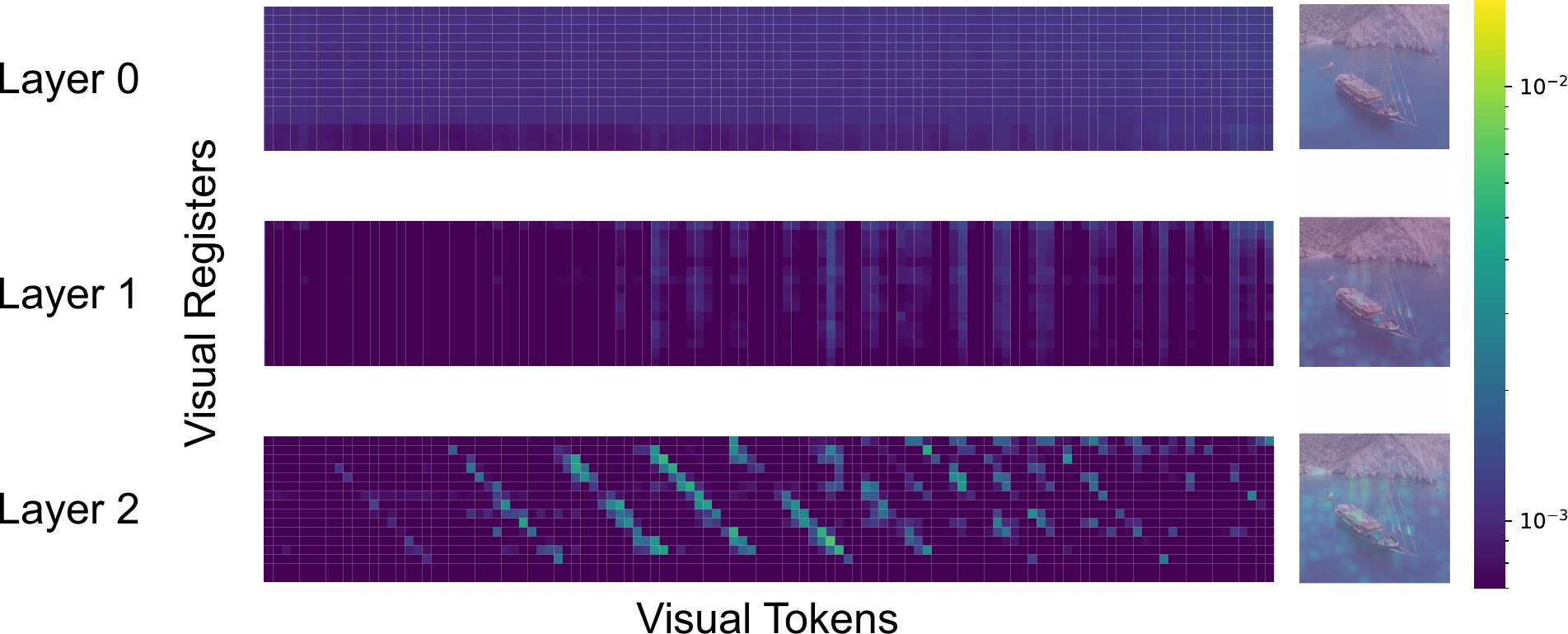}
    \caption{\textbf{Attention Map from Visual Registers to Visual Tokens.} We prompt the model with a test image from the COCO dataset and the instruction, ``Describe the image.''}
    \label{fig:attn_map}
\end{figure}

In \Cref{subsec:motivation}, we empirically demonstrate that the visual registers are more compact than the original visual tokens. In this subsection, we perform a simple analysis to examine whether and how visual registers summarize visual information. The attention map from visual registers to visual tokens is shown in  \Cref{fig:attn_map}. Although the model is not explicitly trained to summarize visual information into the visual registers, they implicitly encode the visual tokens, as indicated by the significant attention scores between visual registers and visual tokens. Interestingly, the visual registers exhibit low attention to visual tokens in the first two layers, and the summarization primarily occurs in the third layer, just before the visual tokens are removed. This may be because the first two layers focus on processing the visual tokens or aligning the visual tokens and registers into a shared space to facilitate communication in later layers. This observation aligns with the ablation results discussed in \Cref{subsec:drop_layer}, where dropping visual tokens in the first or second layer causes a significant performance drop. This suggests that it is more effective for the summarization process to occur in the later layers.

As shown on the right side of \Cref{fig:attn_map}, when examining the attention mapped back to the original image, the visual registers primarily focus on key elements like the rock in the water and the boat mast, while also capturing broader regions of the image. Overall, even without supervision, \texttt{Victor} implicitly learns to summarize the image information both effectively and efficiently.

\section{Limitation and Future Work}

While \texttt{Victor} is simple and effective, we identified some limitations and directions for future improvements. Currently, \texttt{Victor} is not a training-free method, and it must be incorporated at the training stage of the vision-language modeling. Developing a version of \texttt{Victor} that could be applied post-training would be a valuable advancement. However, this might be challenging, as the language tower may need to be specifically trained to learn to effectively utilize the visual registers. Another limitation is the inflexibility of the number of visual registers. As discussed in \Cref{subsec:adjust_num}, the performance degrades when the number of visual tokens is changed on the fly without retraining. In future work, we believe incorporating certain auxiliary loss functions could help make \texttt{Victor} more adaptable and flexible. Additionally, while this paper focuses on applying \texttt{Victor} to vision-language models, we believe this technique could also benefit language models, particularly in long-context tasks. We leave these for future exploration.

\section{Conclusion}

In this paper, we introduce \texttt{Victor}, a novel visual token summarization method that significantly enhances the efficiency-performance trade-off in vision-language models. Without explicit enforcement, the language tower utilizes register tokens to summarize visual information within the first $10\%$ of the layers. After summarization, \texttt{Victor} eliminates the need for visual tokens in the following layers. Our approach offers a superior balance in efficiency compared to state-of-the-art methods. Moreover, with just up to $0.03\%$ additional parameters, \texttt{Victor} is compatible with various attention mechanisms, providing a user-friendly and efficient solution across different hardware environments for future applications.

\section*{Acknowledgements}
We would like to thank Rasoul Shafipour, Seyed-mohsen Moosavi-dezfooli, and Max Horton for their valuable feedbacks and discussions. 

\bibliography{victor}
\bibliographystyle{iclr}

\newpage
\appendix
\section{Appendix}
\subsection{Length Statistics for Individual Benchmarks}
\label{subsec:bench_stats}

We show the length statistics of benchmarks in \Cref{tab:bench_stats}. Based on the representative lengths of these benchmarks, there are two main categories: 1) short-generation, represented by the $2$-token generation scenario in our experiments, and 2) long-generation, represented by the $128$-token generation scenario.

\begin{table}[H]
\tiny
\centering
\caption{\textbf{Prompt and Generation Length Stats of Individual Benchmarks.}}
\text{\textbf{Short-Generation Benchmarks}}
\label{tab:bench_stats}
\begin{tabular}{c|cccccccccc|c}
\toprule
              & VQAv2 & GQA   & ScienceQA & TextVQA & VizWiz & POPE  & MME   & MMBench & Seed  & MMMU   & Average \\
\midrule
Prompt Len. & 43.05 & 46.10 & 93.04     & 43.88   & 44.31  & 43.70 & 54.28 & 86.21   & 93.04 & 210.20 & 75.78   \\
Generation Len. & 1.56  & 1.09  & 2.00      & 8.88    & 3.19   & 1.00  & 1.00  & 1.00    & 2.00  & 1.21   & 2.29  \\
\bottomrule
\end{tabular}
\text{\textbf{Long-Generation Benchmarks}} \\
\begin{tabular}{c|cc|c}
\toprule
              & LLaVA-Bench-Wild & MM-Vet & Average \\
\midrule
Prompt Len. & 49.82            & 49.74  & 49.78   \\
Generation Len. & 146.60           & 96.11  & 121.36 \\
\bottomrule
\end{tabular}
\end{table}

\subsection{Performance on Individual Benchmarks}
\label{subsec:individual_bench}
\begin{figure}[H]
\centering
\begin{subfigure}[t]{0.24\textwidth}
    \includegraphics[width=\textwidth]{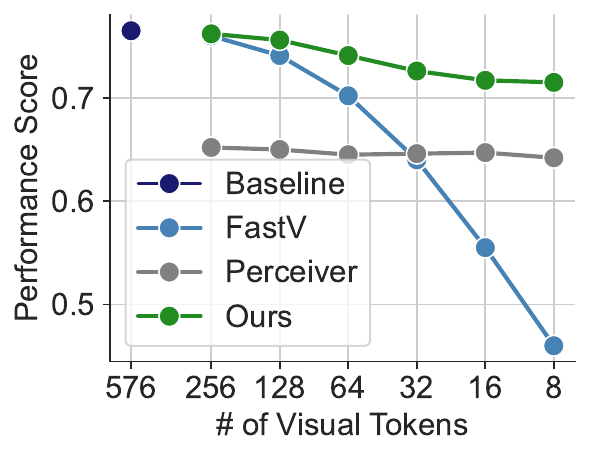}
    \captionsetup{labelformat=empty}
    \caption{VQAv2}
\end{subfigure}
\begin{subfigure}[t]{0.24\textwidth}
    \includegraphics[width=\textwidth]{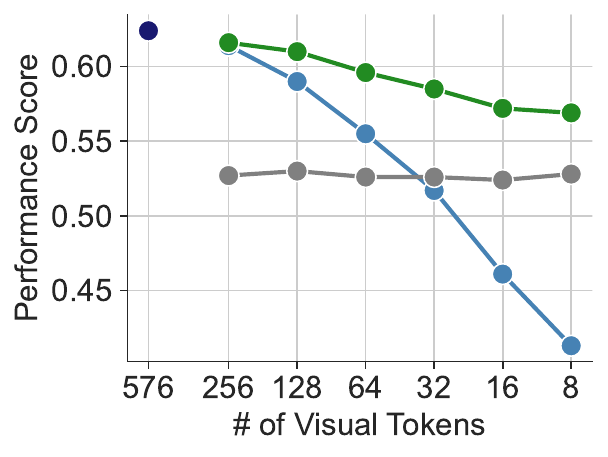}
    \captionsetup{labelformat=empty}
    \caption{GQA}
\end{subfigure}
\begin{subfigure}[t]{0.24\textwidth}
    \includegraphics[width=\textwidth]{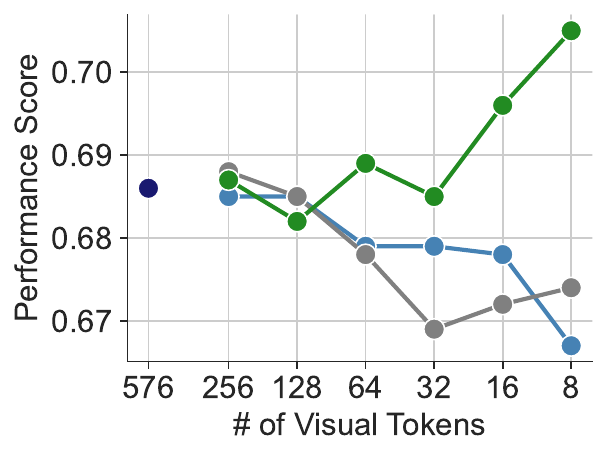}
    \captionsetup{labelformat=empty}
    \caption{ScienceQA}
\end{subfigure}
\begin{subfigure}[t]{0.24\textwidth}
    \includegraphics[width=\textwidth]{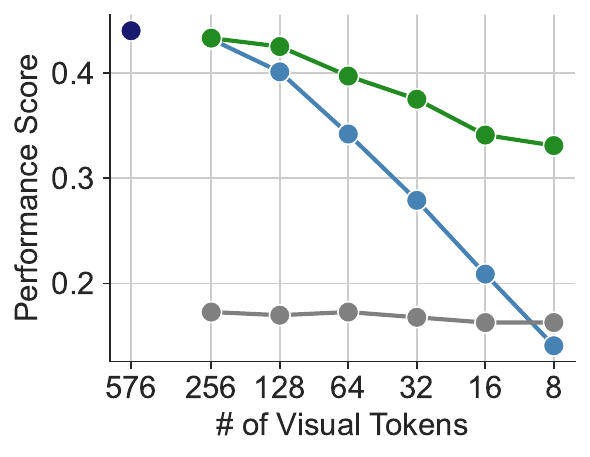}
    \captionsetup{labelformat=empty}
    \caption{TextVQA}
\end{subfigure}
\begin{subfigure}[t]{0.24\textwidth}
    \includegraphics[width=\textwidth]{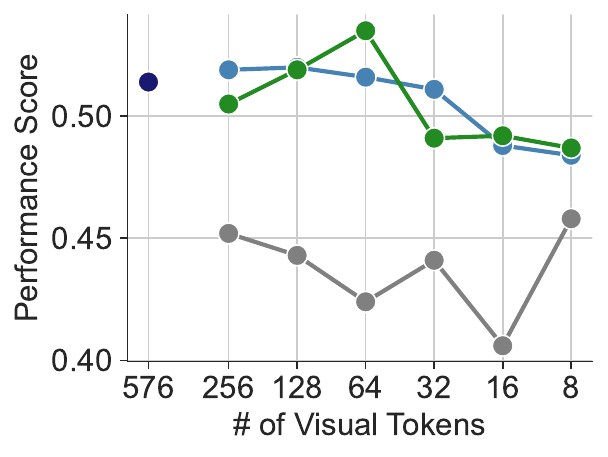}
    \captionsetup{labelformat=empty}
    \caption{VizWiz-VQA}
\end{subfigure}
\begin{subfigure}[t]{0.24\textwidth}
    \includegraphics[width=\textwidth]{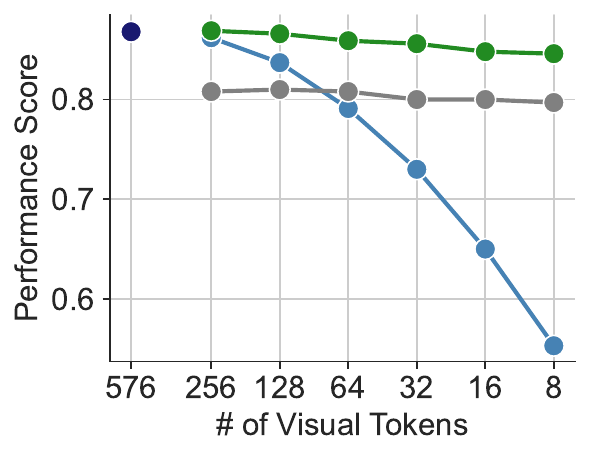}
    \captionsetup{labelformat=empty}
    \caption{POPE}
\end{subfigure}
\begin{subfigure}[t]{0.24\textwidth}
    \includegraphics[width=\textwidth]{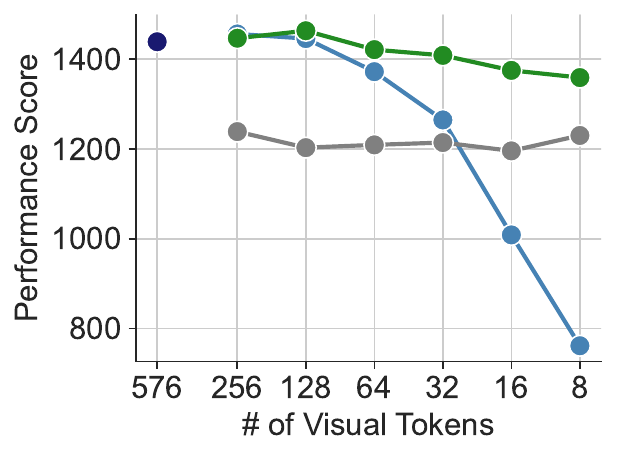}
    \captionsetup{labelformat=empty}
    \caption{MME}
\end{subfigure}
\begin{subfigure}[t]{0.24\textwidth}
    \includegraphics[width=\textwidth]{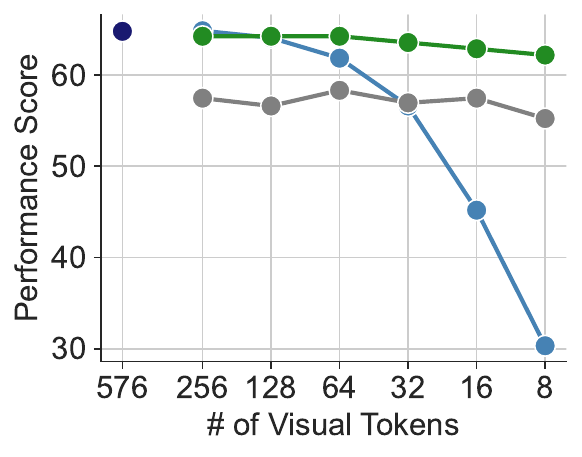}
    \captionsetup{labelformat=empty}
    \caption{MMBench}
\end{subfigure}
\begin{subfigure}[t]{0.24\textwidth}
    \includegraphics[width=\textwidth]{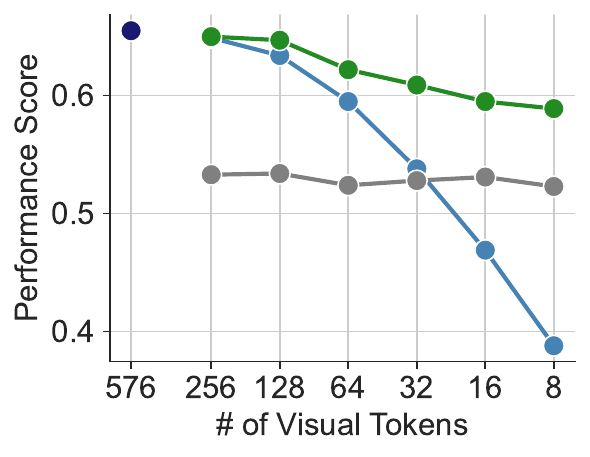}
    \captionsetup{labelformat=empty}
    \caption{SEED-Bench}
\end{subfigure}
\begin{subfigure}[t]{0.24\textwidth}
    \includegraphics[width=\textwidth]{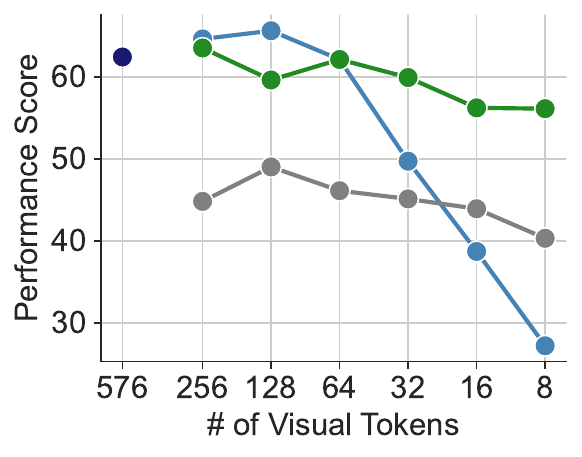}
    \captionsetup{labelformat=empty}
    \caption{LLaVA-Wild}
\end{subfigure}
\begin{subfigure}[t]{0.24\textwidth}
    \includegraphics[width=\textwidth]{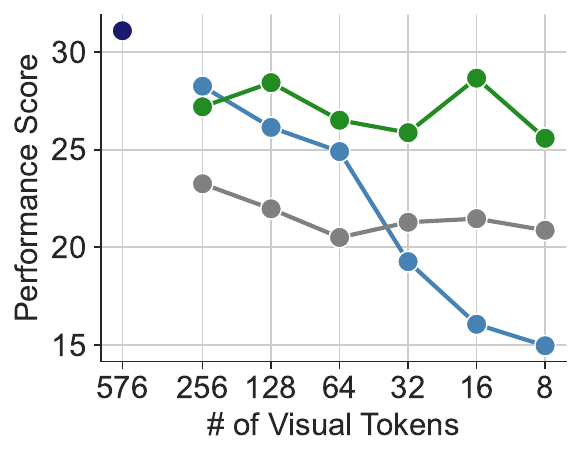}
    \captionsetup{labelformat=empty}
    \caption{MM-Vet}
\end{subfigure}
\begin{subfigure}[t]{0.24\textwidth}
    \includegraphics[width=\textwidth]{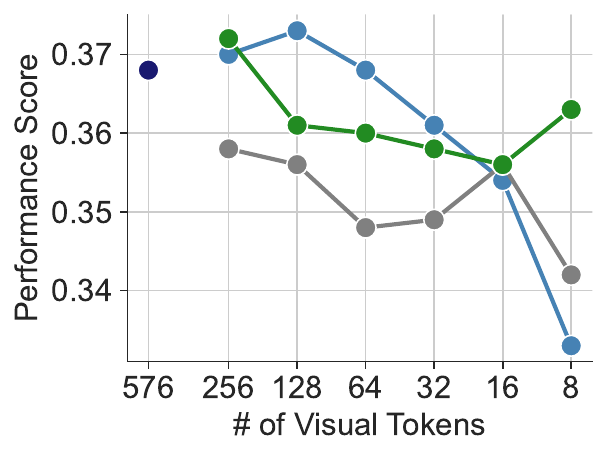}
    \captionsetup{labelformat=empty}
    \caption{MMMU}
\end{subfigure}
\caption{\textbf{Individual Benchmark Performance.}}
\label{fig:individual_performance}
\end{figure}

We show the performance on individual benchmarks of \Cref{subsec:main_result} in \Cref{fig:individual_performance}.

\subsection{Total Training-Time Reduction}
\label{subsec:total_time}
\begin{figure}[H]
    \centering
    \includegraphics[width=0.49\textwidth]{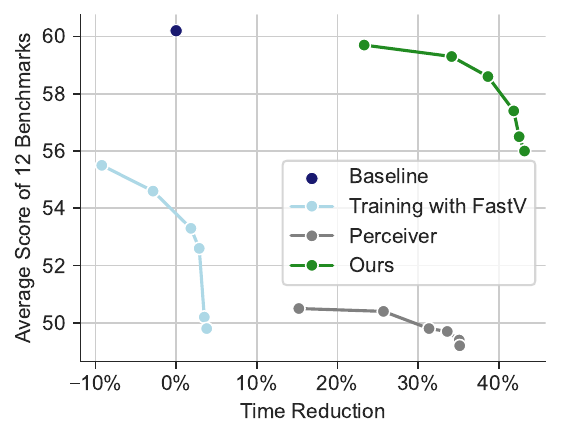}
    \caption{\textbf{Performance vs. Total Training-Time Reduction..}}
    \label{fig:total_time}
\end{figure}

The total training-time reduction is shown in \Cref{fig:total_time}.

\subsection{Extra Results with Different Language Towers}
\label{subsect:app_diff_models}
\begin{figure}[H]
\centering
\begin{subfigure}[t]{0.49\textwidth}
    \includegraphics[width=\textwidth]{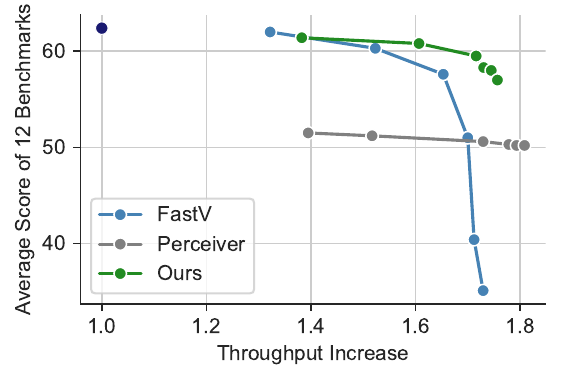}
    \caption{Vicuna-13B-v1.5}
\end{subfigure}
\hfill
\begin{subfigure}[t]{0.49\textwidth}
    \includegraphics[width=\textwidth]{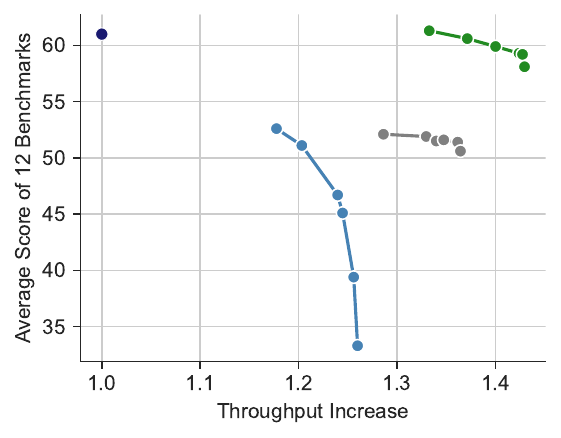}
    \caption{Meta-Llama-3-8B-Instruct}
\end{subfigure}
\hfill
\begin{subfigure}[t]{0.49\textwidth}
    \includegraphics[width=\textwidth]{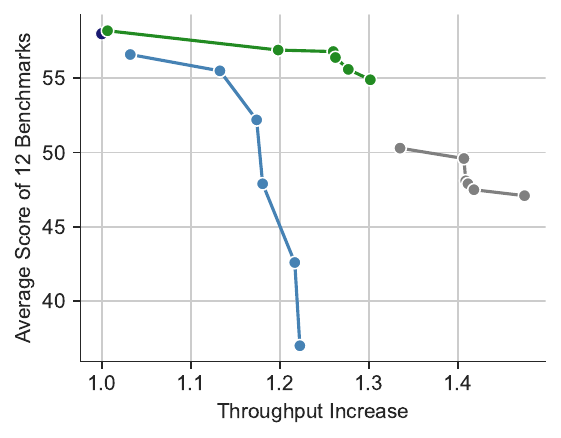}
    \caption{Mistral-7B-Instruct-v0.2}
\end{subfigure}
\hfill
\begin{subfigure}[t]{0.49\textwidth}
    \includegraphics[width=\textwidth]{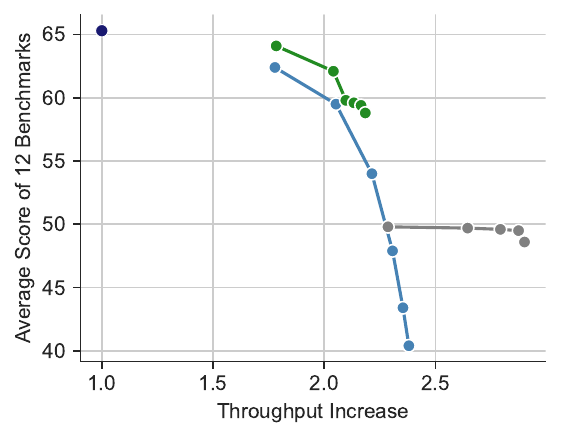}
    \caption{LLaVA-NeXT with Qwen2-7B-Instruct}
\end{subfigure}
\caption{\textbf{Efficiency-Performance Trade-Off Curve with Different Language Towers under $128$-Token Generation.}}
\label{app:fig:s1_diff_model}
\end{figure}

The extra result of \Cref{subsec:diff_model} with $128$-token generation scenario is presented in \Cref{app:fig:s1_diff_model}.

\subsection{Ablation on Adjusting the Number of Visual Registers at Inference}
\label{subsec:adjust_num}

\begin{figure}[H]
    \centering
    \includegraphics[width=0.49\textwidth]{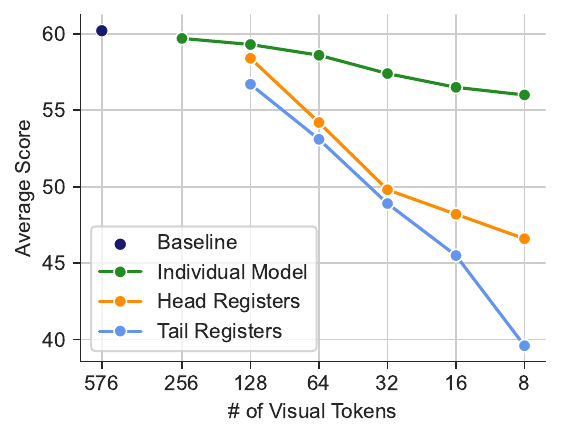}
    \caption{\textbf{Ablation on Adjusting the Number of Visual Registers at Inference.}}
    \label{fig:adjust_num}
\end{figure}

In our main experiments, we retrain the model whenever a different number of visual registers is required. In this subsection, we explore two strategies for dynamically adjusting the number of visual registers at inference time. Given a \texttt{Victor} model with $M$ visual registers, if we want to use $M' < M$ registers, we either select the first $M'$ registers (referred to as ``head'') or the last $M'$ registers (referred to as ``tail''). As shown in \Cref{fig:adjust_num}, the performance of these adjustments is not as effective as retraining the model from scratch. However, we believe adding certain auxiliary losses during training can make our method more flexible, and we leave this for future work.

\subsection{Importance of Visual Registers for Summarization}
\begin{figure}[H]
    \centering
    \includegraphics[width=0.49\textwidth]{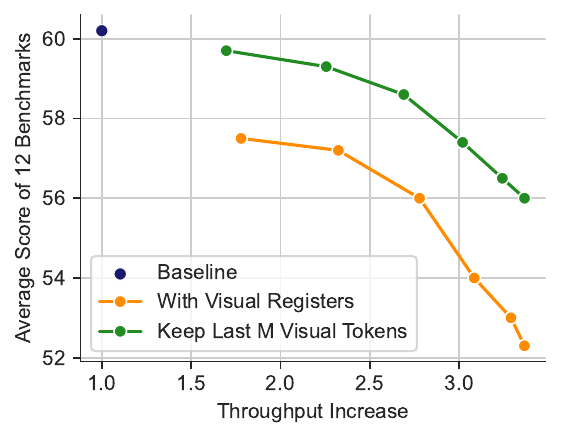}
    \caption{\textbf{Importance of Visual Registers for Summarization.}}
    \label{fig:no_register}
\end{figure}

In this subsection, we conduct an ablation study to demonstrate the necessity of using visual registers for summarization. Specifically, we compare our approach to an alternative method where instead of prepending additional tokens to the visual tokens, we retain the last $M$ visual tokens at layer $3$. This requires the model to summarize all visual information into these last existing $M$ visual tokens. As shown in \Cref{fig:no_register}, while the ablated method results in a slight improvement in throughput, overall the performance drops significantly. This highlights the importance of incorporating visual registers for effective summarization.

\subsection{Different Visual Registers}
\begin{figure}[H]
  \begin{center}
    \includegraphics[width=0.49\textwidth]{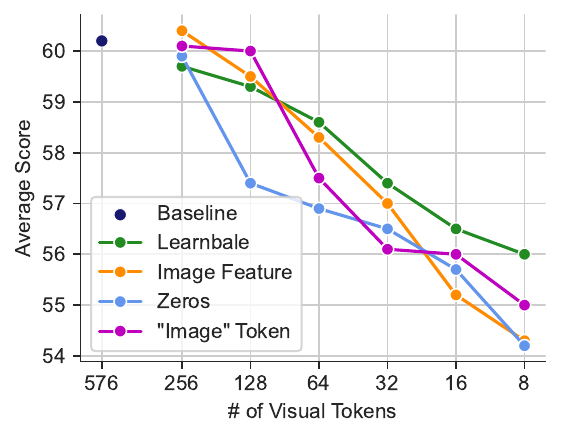}
  \end{center}
  \caption{\textbf{Ablation with Different Visual Registers.}}
  \label{fig:diff_init}
\end{figure}

We also experiment with various types of visual registers. In addition to using learnable tokens, we test three alternative methods for visual registers: 1) \textbf{Pooled Image Feature}: utilizing average-pooled visual tokens as the register tokens, 2) \textbf{Zeros}: initializing with all zeros, and 3) \textbf{``Image'' Token}: using the embedding of the word ``Image.'' The results are presented in \Cref{fig:diff_init}. The ``Image'' Token method is effective for the visual registers, especially when the number of visual tokens is reduced to $256$ and $128$, as there is no performance drop. However, all alternative methods showed relatively worse performance compared to learnable queries in the low visual token regime. Therefore, we adopt learnable queries for \texttt{Victor} as they offer better overall performance.

\end{document}